\documentclass[letterpaper, 10 pt, conference]{ieeeconf}
\IEEEoverridecommandlockouts
\usepackage{cite}
\usepackage{amsmath,amssymb,amsfonts}
\usepackage{algorithmic}
\usepackage{graphicx}
\usepackage{subfigure}
\usepackage{textcomp}
\usepackage{xcolor}
\usepackage[ruled]{algorithm2e}
\usepackage{booktabs}
\usepackage{multirow}
\usepackage{bbm}

\usepackage{soul}

\def\BibTeX{{\rm B\kern-.05em{\sc i\kern-.025em b}\kern-.08em
    T\kern-.1667em\lower.7ex\hbox{E}\kern-.125emX}}

\title{\LARGE \bf
Reward-Driven Automated Curriculum Learning for Interaction-Aware Self-Driving at Unsignalized Intersections
}

\author{Zengqi Peng, Xiao Zhou, Lei Zheng, Yubin Wang, and Jun Ma
\thanks{This work was supported in part by the National Natural Science Foundation of China under Grant 62303390; in part by the Guangzhou-HKUST(GZ) Joint Funding Scheme under Grants 2023A03J0148 and 2024A03J0618; and in part by the Project of Hetao Shenzhen-Hong Kong Science and Technology Innovation Cooperation Zone under Grant HZQB-KCZYB-2020083. \textit{(Corresponding Author: Jun Ma.)}}
\thanks{Zengqi Peng, Xiao Zhou, Lei Zheng, and Yubin Wang are with the Robotics and Autonomous Systems Thrust, The Hong Kong University of Science and Technology
(Guangzhou), Guangzhou, China
(e-mail: zpeng940@connect.ust.hk; xzhou910@connect.hkust-gz.edu.cn; lzheng135@connect.ust.hk; ywang575@connect.hkust-gz.edu.cn).
}
        \thanks{Jun Ma is with the Robotics and Autonomous Systems Thrust, The Hong Kong University of Science and Technology (Guangzhou), Guangzhou, China, also with the Division of Emerging Interdisciplinary Areas, The Hong Kong University of Science and Technology, Hong Kong SAR, China, and also with  HKUST Shenzhen-Hong Kong Collaborative Innovation Research Institute, Futian, Shenzhen, China (e-mail: jun.ma@ust.hk).}
}

\begin{document}

\maketitle
\thispagestyle{empty}
\pagestyle{empty}

\begin{abstract}
In this work, we present a reward-driven automated curriculum reinforcement learning approach for interaction-aware self-driving at unsignalized intersections, taking into account the uncertainties associated with surrounding vehicles (SVs). These uncertainties encompass the uncertainty of SVs' driving intention and also the quantity of SVs. To deal with this problem, the curriculum set is specifically designed to accommodate a progressively increasing number of SVs. By implementing an automated curriculum selection mechanism, the importance weights are rationally allocated across various curricula, thereby facilitating improved sample efficiency and training outcomes. Furthermore, the reward function is meticulously designed to guide the agent towards effective policy exploration. Thus the proposed framework could proactively address the above uncertainties at unsignalized intersections by employing the automated curriculum learning technique that progressively increases task difficulty, and this ensures safe self-driving through effective interaction with SVs. Comparative experiments are conducted in $Highway\_Env$, and the results indicate that our approach achieves the highest task success rate, attains strong robustness to initialization parameters of the curriculum selection module, and exhibits superior adaptability to diverse situational configurations at unsignalized intersections. Furthermore, the effectiveness of the proposed method is validated using the high-fidelity CARLA simulator.

\end{abstract}

\section{Introduction}

In recent years, autonomous driving technologies have achieved notable advancements and remarkable strides in both academic and industrial domains \cite{paden2016survey,muhammad2020deep,grigorescu2020survey}. Nevertheless, self-driving tasks in dense and interaction-heavy scenarios remain an open challenge, 
primarily stemming from the lack of precise driving intention of surrounding vehicles (SVs) 
\cite{mozaffari2020deep,li2021open}. 
Apparently, inaccurate assessments regarding the driving intention of SVs could significantly hinder safe motion planning and potentially lead to traffic accidents \cite{zheng2023spatiotemporal}. 
The situation becomes even more severe when it comes to unsignalized intersections, where the ego vehicle (EV) needs to coordinate simultaneously with multiple vehicles approaching from various directions \cite{wei2021autonomous}. 
In general, these intersections exhibit unpredictable traffic patterns, and uncertainties in the driving behavior of SVs pose serious safety concerns during the interaction process, making it challenging for autonomous vehicles to anticipate and respond appropriately to potential hazards. In this sense, unsignalized intersections typically require complex decision-making and situational awareness.

Rule-based and optimization-based methods are extensively studied towards the decision-making and planning of autonomous vehicles at intersections. 
Rule-based methods can prioritize driving safety and avoid potential collisions among road users by designing a set of rules in accordance with road traffic safety regulations \cite{aksjonov2021rule}. 
However, rule-based methods struggle to account for all possible traffic scenarios and tend to be overly cautious, leading to an excessive level of conservatism \cite{tian2020game}. 
On the other hand, optimization-based methods enable the EV to navigate at intersections by minimizing a given objective function subject to well-designed constraints \cite{riegger2016centralized,kneissl2018feasible}. However, these methods are computationally expensive and time-consuming, and they may not always be able to account for unexpected or rapidly changing traffic scenarios, making them less adaptable in certain situations. Additionally, optimization-based methods are prone to getting stuck in the local optima when it comes to complex situations (e.g., scenarios involving SVs with interactive driving behaviors), which could lead to suboptimal or even unsafe decisions being made by the autonomous vehicle.

Reinforcement learning (RL) 
has recently demonstrated tremendous potential in autonomous driving \cite{isele2018navigating,kiran2021deep,xia2022interactive,qiao2021behavior}. 
Nevertheless, one significant limitation of RL-based approaches is the low sample efficiency when dealing with complex driving scenarios \cite{huang2023efficient}. The agent requires a significant amount of time exploring the environment due to the lack of appropriate guidance. 
Furthermore, training directly on the data from complex target tasks can potentially lead the network to converge to a locally optimal policy, resulting in poor driving performance and adaptivity capability. One promising way to alleviate the above problem is curriculum learning \cite{bengio2009curriculum,song2021autonomous,wang2023chance}. 
In \cite{peng2023CPPO}, a curriculum reinforcement learning approach is proposed for self-driving tasks. However, as the timing of curricular transitions is manually set, the quality of training outcomes will be highly dependent on the expert experience. Besides, the parameters governing the timing of curriculum transitions exhibit substantial variation across different scenarios.

To address the above problem, a series of automated curriculum learning methods have been proposed \cite{graves2017automated,qiao2018automatically,khaitan2022state}. 
In \cite{graves2017automated}, two learning progress signals are designed for the automated transition of curriculum for neural networks. 
Particularly, the impact of two signals on the training process is compared across different curricula. 
An action-value-based RL approach is proposed in \cite{qiao2018automatically} to address the autonomous driving task at single-lane crossroads with stop signs by modeling different starting positions of the EV as a curriculum set. 
Nonetheless, the update of curriculum importance is solely related to their respective weights, 
potentially leading to curricula with high future rewards being overlooked. 
In \cite{khaitan2022state}, a state drop-out framework is designed to train agents for traversing the unsignalized intersection. The agent can obtain a fine-tuned policy in the final phase by dropping the future state information during the training process. However, SVs are set as non-interactive, and their future trajectories are assumed to be accessible to the EV in \cite{qiao2018automatically} and \cite{khaitan2022state}. These assumptions could compromise driving safety and limit the adaptivity of the agent when it comes to realistic driving situations. 
It is pertinent to note that, in real-world driving situations, drivers display different levels of interactive behaviors with SVs, which is a crucial factor to consider for ensuring safe interactions. 
Essentially, most existing works do not adequately address the interaction behaviors of SVs, which could restrict the applicability of the trained policy in real-world settings.

\begin{figure}[!t] 
    \centering   
    \includegraphics[trim=0 0.7cm 0 0.5cm, width=0.95\linewidth]{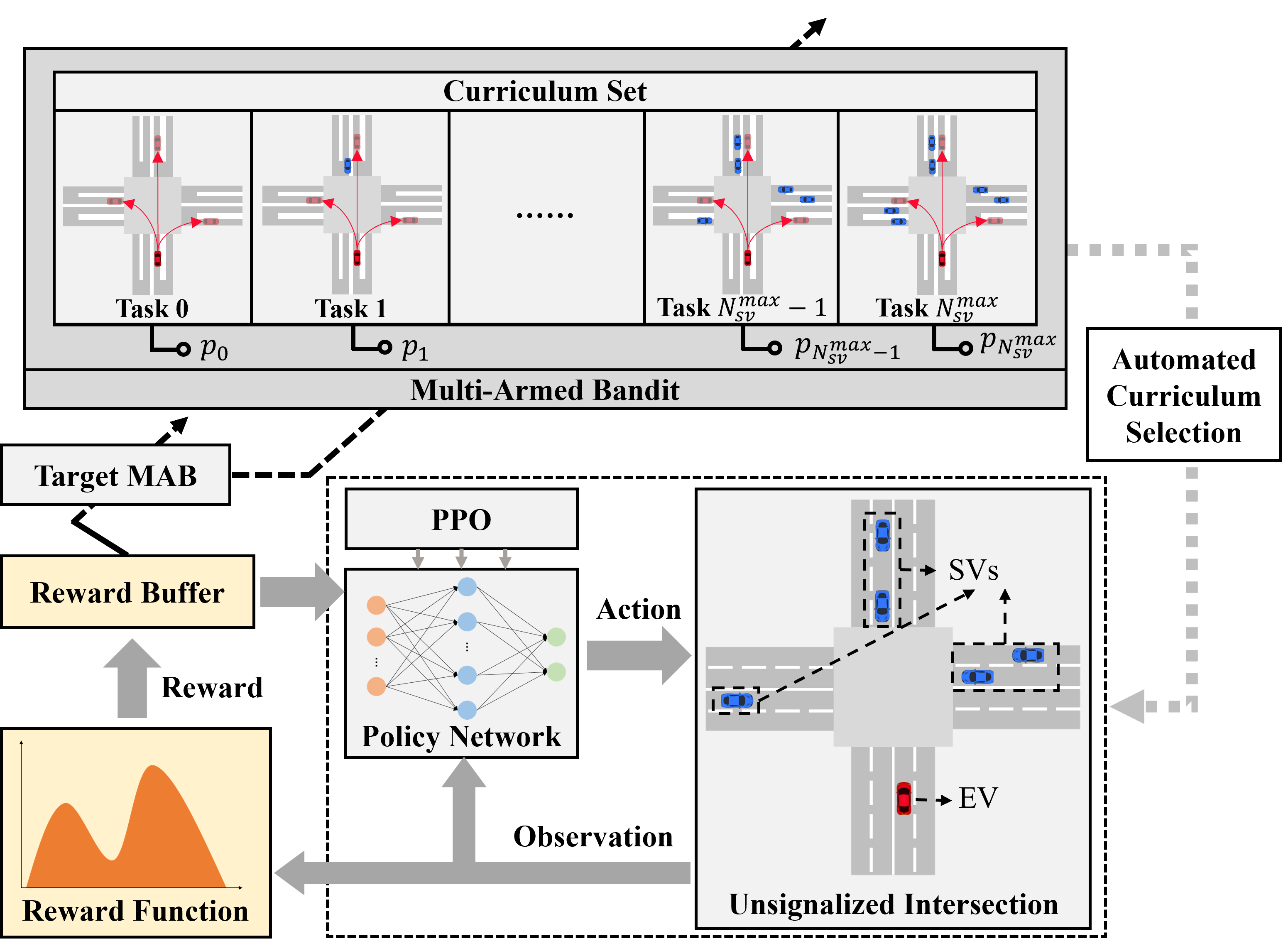}
   \caption{Overview of the proposed framework for autonomous driving at unsignalized intersections with interactive SVs. In the four-way intersection scenario, the EV is depicted in red, and the SVs are in blue. The solid vehicle and the semi-transparent vehicle represent the start point and the goal point, respectively. 
   }
    \label{frame}
\end{figure}

To address the above challenges, we present a reward-driven automated curriculum proximal policy optimization (RD-ACPPO) approach for self-driving tasks at unsignalized intersections. 
The pipeline of the proposed framework is shown in Fig. \ref{frame}. 
Particularly, we decompose the target task into a sequence of generalized curricula with progressively increasing difficulty levels and model the self-driving tasks as a multi-armed bandit (MAB) problem.
An exponential-weight-based curriculum selection module is devised to evaluate the importance weights of arms of MAB during the training, which is then integrated with the proximal policy optimization (PPO) framework to facilitate efficient training for the RL policy.
The main contributions are summarized as follows:

(1) An interaction-aware RD-ACPPO framework is proposed for self-driving tasks at unsignalized intersections. 
Particularly, an adaptive curriculum learning technique is presented to accommodate a progressively increasing difficulty of learning tasks in autonomous driving. 
In this sense, the proposed approach is able to
proactively account for the uncertainty arising from the intention of SVs and the traffic density while adequately addressing the interaction behaviors of SVs. 

(2) A distinctive automated curriculum selection module is devised for efficient sampling of episodes in the learning process. By modeling the selection process of curricula as an MAB problem, the importance weights of curricula are dynamically adjusted by real-time assessments of the training progress, which facilitates efficient and effective exploration within the environment for the RL agent.

(3) The effectiveness of the proposed framework is demonstrated in simulated unsignalized intersection environments, with comparisons conducted against baseline methods. The results show that the proposed approach exhibits appropriate curriculum transition timing and strong robustness to initialization parameters of MAB, and its trained policy attains superior adaptivity to diverse situational configurations with the highest task success rate among all methods.

\section{Problem Definition}
\subsection{Problem Statement}

The task scenario is set as an unsignalized four-way intersection. 
The task type, including left turns, going straight, and right turns, is randomly assigned to the EV and SVs. 
The EV always starts from a random position within the lower lane, and its goal position is also generated randomly within the other three roads, obeying the traffic rules. The SVs are generated on the other three roads randomly, and they drive towards different target lanes. 

Here, we assume that the position and velocity information of SVs can be accessed by the EV while the information about their driving intention is unknown. 
In contrast with \cite{qiao2018automatically} and \cite{khaitan2022state}, each SV is set to interact with other vehicles in our task scenario, and the number of SVs is not fixed in different tasks. Apparently, the settings in our task scenario introduce noticeable randomness into the task process, which makes the problem challenging but close to real-world conditions. 
The objective is to generate a sequence of actions to enable the EV to efficiently approach the target point, avoid collisions with other SVs, and stay within road boundaries.

\subsection{Learning Environment}

Typically, the self-driving task at unsignalized intersections is modeled as a Markov Decision Process (MDP) \cite{peng2023CPPO}. It can be expressed as a tuple $\mathcal{E} = \langle \mathcal{S}, \mathcal{A}, \mathcal{P}, \mathcal{R}, \gamma \rangle$, where $\mathcal{S}, \mathcal{A}, \mathcal{P}, \mathcal{R}, \gamma$ represent the state space, action space, state transition dynamics, reward function, and discount factor, respectively. 

\textbf{State space $\mathcal{S}$}: In our problem, the state space $\mathcal{S}$ consists of kinematic features of all vehicles. At time step $t_s$, the state matrix is defined as:
\begin{equation}
\begin{split}
\mathbf{S}_{t_s}=\left[\ \mathbf{s}_{t_s}^0\ \ \mathbf{s}_{t_s}^1\ ...\ \mathbf{s}_{t_s}^{N_{sv}^{max}}\ \right]^T,
\end{split}
\label{state martix}
\end{equation}
where $N_{sv}^{max}$ denotes the maximum number of SVs; $\mathbf{s}_{t_s}^0$ is the state of the EV; $\mathbf{s}_{t_s}^i\ (i=1,2,...,N_{sv}^{max})$ represents the state of the $i$-th SV, which is defined as follows:
\begin{equation}
\begin{split}
\mathbf{s}_{t_s}^{i}={\left[\begin{array}{l l l l l l l}{x_{t_s}^{i}}&{y_{t_s}^{i}}&{v_{x,t_s}^{i}}&{v_{y,t_s}^{i}}&{\sin\psi_{t_s}^{i}}&{\cos\psi_{t_s}^{i}}\end{array}\right]}^{T},
\end{split}
\label{kinematic_features}
\end{equation}
where $x_{t_s}^{i}$ and $y_{t_s}^{i}$ represent the X-axis and Y-axis coordinates of the vehicle $i$ in the world coordinate system, respectively; $v_{x,t_s}^{i}$ and $v_{y,t_s}^{i}$ are the speed of the $i$-th vehicle along the corresponding axes, respectively; $\psi_{t_s}^{i}$ is the heading angle of the $i$-th SV. 

\textbf{Action space $\mathcal{A}$}: In our MDP, a discrete action space consisting of five actions is utilized for the RL agent, which is defined as follows:
\begin{equation}
\begin{split}
\mathcal{A}=\left\{ a_{1},a_{2},a_{3},a_{4},a_{5} \right\},
\end{split}
\label{Action space}
\end{equation}
where $a_{1}, a_{2}, a_{3}, a_{4}$, and $a_{5}$ represent the left lane-changing, motion-keeping, right lane-changing, deceleration, and acceleration action, respectively.
The RL agent needs to choose a high-level action from the action space $\mathcal{A}$. Then the selected action is converted into continuous control signals for the vehicle by low-level controllers.

\textbf{State transition dynamics $\mathcal{P}(\mathbf{S}_{t_s+1}|\mathbf{S}_{t_s},a_{t_s})$}: The transition function defines the transition of state matrix $S_{t_s}$, which follows the Markov transition distribution. It is implicitly defined by the environment and unknown to the agent.

\textbf{Reward function $\mathcal{R}$}: This work assigns a positive reward when the agent successfully completes a self-driving task or maintains survival during the task process. It penalizes collisions, out-of-the-road, and frequent lane-changing behaviors. The details about the reward function will be discussed in Section III-C.

\textbf{Discount factor $\gamma$}: The future reward is accumulated with a discount factor $\gamma \in (0,1)$.

\section{Methodology}

\subsection{Task Decomposition and Curriculum Modelling}
In this work, we consider the training process of the RL policy as a curriculum learning problem with the target task setting, containing $N_{sv}^{max}+1$ curriculums with an increasing number of SVs. 
The curriculum set is defined as 
\begin{equation}
\begin{split}
\mathbf{\Omega}=\left\{\Omega_i | i = 0, 1, 2, ..., N_{sv}^{max} \right\},
\end{split}
\label{Curr_Set}
\end{equation}
where $i$ represents the serial number of the task, which is set to be equivalent to the number of SVs $N_{sv}$ in the particular curriculum. 
For each curriculum in this set, we can consider it as an arm in the MAB problem \cite{slivkins2019introduction}. 
When we select a curriculum to collect an episode, we can obtain related states and rewards from the environment. This process is similar to the one-step run in the MAB, where one arm is chosen and then a reward is obtained. 
Based on the above analysis, we consider the problem of the curriculum selection in (\ref{Curr_Set}) as a sampling process of the MAB $\mathcal{M}$ with $N_{sv}^{max}+1$ arms. 
Under this setting, the MAB agent will sample a sequence of arms over $T$ rounds during the training process:
\begin{equation}
 A_T = \left\{ a_{s,1}, a_{s,2}, ..., a_{s,T} \right\},
\label{MAB_arm_sample}
\end{equation}
where $a_{s,t}\ (t=1,2,...,T)$ is the MAB arm sampled in the $t$-th round. 
After each round, the selected arm returns a reward, and the MAB agent updates the importance weights according to the historical reward. 
We aim to obtain an adaptive mechanism that can maximize payoffs obtained by the sampled curriculum sequence $A_T$ from the MAB $\mathcal{M}$:
\begin{equation}
\mathbf{w}^*=\arg \max _{\mathbf{w},A_T} \sum\nolimits_{t=1}^T \hat{r}(t),
\label{Equ:MAB_max}
\end{equation}
where $\mathbf{w}$ is the importance weight vector of the MAB; $\hat{r}(t)$ denotes the rescaled reward, which is related to the received reward of the RL agent. 

\subsection{Automated Curriculum Selection}
For the MAB model constructed in Section III-A, we define its importance weight vector $\mathbf{w}(t)$ and probability distribution vector $\mathbf{p}(t)$ as follows:
\begin{equation}
\begin{array}{l}
\mathbf{w}(t)=\left\{\mathrm{w}_i(t) | i = 0, 1, ..., N_{sv}^{max} \right\},\\
\mathbf{p}(t)=\left\{ p_i(t) | i = 0, 1, ..., N_{sv}^{max} \right\}.
\end{array}
\label{def:weight_pobs}
\end{equation}

In our specific problem, the optimal arm fluctuates throughout different training stages due to shifts in the expected rewards, which is associated with specific tasks. 
Inspired by the Exp3 algorithm \cite{auer2002nonstochastic}, this problem can be addressed by incorporating an external $\varepsilon$-greedy term into the update of 
the probability. This guarantees that all arms have a probability of being selected throughout the entire training process. 
Given the weight vector $\mathbf{w}(t)$ at $t$-th episode, we can calculate the sampling probability of the arm $i$ by the following exponential-weight algorithm:
\begin{equation}
\begin{split}
p_i(t)=(1-\eta) \frac{e^{\mathrm{w}_i(t)}}{\sum_{j=0}^{N_{sv}^{max}} e^{\mathrm{w}_j(t)}}+\frac{\eta}{N_{sv}^{max}+1}, \\ i=0, 1, ..., N_{sv}^{max},
\end{split}
\label{equ:pobs_exp_cal}
\end{equation}
where $\eta \in (0,1)$ is a constant parameter that mediates the trade-off between the exploitation of experience data and random exploration. 
Then we derive a curricular sample $\Omega(t)$ at $t$-th episode from the calculated distribution $\mathbf{p}(t)$ as follows:
\begin{equation}
    \Omega(t) \sim \mathcal{M}(\mathbf{p}(t)).
\label{equ:MAB_sample}
\end{equation}

The MAB agent will receive a reward $r_i(t)$ from the training episode. Then the reward is rescaled to evaluate the performance of the selected arm as follows:
\begin{equation}
\left\{
\begin{array}{l}
\hat{r}_i(t)=\frac{r^i_{norm}(t)}{p_i(t)}, \\
r^i_{norm}(t)=\frac{2\left(r_i(t) - k_0 R_{min}(t)\right)}{k_1 R_{max}(t) - k_0 R_{min}(t)}-1,
\end{array}
\right.
\label{Exp3:rescale}
\end{equation}
where $\hat{r}_i(t)$ is the rescaled reward of $i$-th arm obtained in the $t$-th episode; $R_{max}(t)$ and $R_{min}(t)$ are the maximum reward and minimum reward in the history of unscaled rewards up to $t$-th episode, respectively; $k_0$ and $k_1$ are two constant parameters to adjust the rescaling process. Here, we divide all obtained rewards by their current probability, ensuring that arms with potential optimality but currently low probability can be promptly discovered with policy updates. 
Then we update the weight vector of $i$-th arm as follows:
\begin{equation}
\mathrm{w}_i(t+1)=\mathrm{w}_i(t) + \alpha \hat{r}_i(t),
\label{equ:weight_update}
\end{equation}
where $\alpha \in (0,1)$ is a constant parameter for adjusting the growth rate of importance weights of each arm.

Considering the nature of the problem addressed in this work, the complexity and difficulty of task scenarios within our task set increase progressively. 
As the training curriculum is sampled via MAB, and the positions of SVs are randomly generated, low efficient samples are difficult to completely avoid during the training process. 
To mitigate the overestimation or underestimation in the reward of arms due to occasional sampling, 
we introduce a target MAB $\hat{\mathcal{M}}$ for stable updating of MAB. After random sampling in the MAB, the obtained rescaled rewards are used to update the parameters of the target MAB. The parameters of the MAB are synchronized with the target MAB after a certain amount of samplings. 
Therefore, the update of the weight of arm $i$ can be expressed as follows:
\begin{equation}
\mathrm{w}_i(t+1)=\left\{
\begin{array}{l}
\mathrm{w}_i(t),\ \text{if}\ \ t | N_{MAB} \neq 0, \\
\mathrm{w}_i(t) + \sum_{j=0}^{N_{MAB}-1} \alpha \hat{r}_i(t-j) ,\ \text{otherwise},
\end{array}
\right.
\label{equ:weight_double}
\end{equation}
where $N_{MAB}$ is the update interval for synchronizing the parameters of target $\hat{\mathcal{M}}$ to $\mathcal{M}$.

\subsection{Reward-Driven Automated Curriculum Proximal Policy Optimization}
Based on the automated curriculum selection module, a specific arm is sampled from the MAB to generate an episode for the training of the RL policy. Corresponding information including observations, actions, and rewards is stored in a replay buffer for offline training. The RL policy is trained to maximize the following cumulative objective function related to the sampled curriculum sequence $A_T$ as follows:
\begin{equation}
\theta^*=\arg \max _{\theta,A_T} J(\theta_t),
\label{equ:ACRL}
\end{equation}
where $J(\theta_t)$ is the objective function for policy network with $\theta_t$. 
In this study, we adopt the clipped objective function of PPO algorithm \cite{schulman2017proximal} for training the RL policy:
\begin{equation}
J(\theta_t)=\mathbb{E}_t\left[\min \left(\rho_t(\theta_t) \hat{A}_t, \operatorname{clip}\left(\rho_t(\theta_t), 1-\epsilon, 1+\epsilon\right) \hat{A}_t\right)\right],
\end{equation}
where $\rho_t(\theta)$ represent the similarity between the new policy and old policy; $\hat{A}_t$ is the estimated advantage function; $\epsilon$ is the clip parameter.

The reward function plays a pivotal role in guiding RL agents to explore the environment. 
In this work, the reward function is particularly designed for interactive task scenarios with varying numbers of vehicles. 
Considering the characteristics of autonomous driving tasks at unsignalized intersections, the reward function is designed as follows:
\begin{equation}
\begin{split}
    r(t) &=  r_{S}(I_S) + r_{C}(I_C) + r_{TO}(I_{TO})\\ 
    & \quad + r_{OfR}(I_{OfR}) + r_{LC}(I_{LC}) + r_{l},
\end{split}
\label{reward_func}
\end{equation}
where $r_{S}$ and $r_{l}$ are the reward for successfully completing tasks and surviving in the task, respectively; $r_{C}$, $r_{TO}$, $r_{OfR}$, and $r_{LC}$ are the penalty of collision with SVs, time-out, out of the road boundary, and lane-changing behavior, respectively. $I_{event}$ is the indicator function corresponding to the occurrence of different events as follows: 
\begin{equation*}
I_{event}=\left\{
\begin{array}{l}
1,\ \text{if event occurs}, \\
0,\ \text{otherwise}.
\end{array}
\right.
\end{equation*}

To achieve high task efficiency, we set the success reward term $r_{S}$ to be dependent on both the task completion time $t_c$ and the number of SVs $N_{sv}$ when passing through the intersection. This configuration can incentivize RL agents to explore complex task scenarios. 
Furthermore, to encourage the RL agent to obtain and enhance the interaction-aware and collision-avoidance capability, the collision penalty $r_{C}$ is set to be positively correlated with the speed of the EV $v_{EV}$ and the number of SVs when a collision occurs. 
Additionally, the lane-changing penalty is positively correlated with the number of lane changes $n_{lc}$ during the task to avoid frequent lane-changing behavior of the RL agent.
$r_{S}$ and $r_{C}$ are set as follows:
\begin{equation}
\begin{aligned}
r_{S}(I_S) &= I_S \cdot (\alpha_1 \cdot \frac{t_c}{t_c^{max}} \cdot N_{sv}^2 + \alpha_2),\\
r_{C}(I_C) &= I_C \cdot (\alpha_3 \cdot v_{EV} \cdot N_{sv}^2 + \alpha_4),
\end{aligned}
\label{def:weight_pobs}
\end{equation}
where $\alpha_i,i=1,2,3,4$ are constant parameters.
The remaining reward terms are set to constants if the corresponding events occur.

The complete procedure of the proposed RD-ACPPO framework is summarized in \textbf{Algorithm \ref{ALG_CPPO}}.

\begin{algorithm}[t]  
	\caption{RD-ACPPO} 
        \label{ALG_CPPO}
	\LinesNumbered 
	\KwIn{Environmental state $\mathbf{S}_{t_s}$, curriculum set $\mathbf{\Omega}$, MAB update frequency $N_{MAB}$}
	\KwOut{$\pi^* = f(\mathbf{\theta}^*)$}
        Initialize the MAB and target MAB with arm's weights $\left\{ \mathrm{w}_i \right\}$ and $\left\{ \hat{\mathrm{w}}_i \right\}$, where $i=0,1,...,N_{sv}^{max}$, 
        the policy network with parameter $\mathbf{\theta}_0$\; 
        \While{$t \leq t_{max}$}  
            {Calculate the probability distribution $\mathbf{p}(t)$ according to (\ref{equ:pobs_exp_cal})\;
            Derive a sample $\Omega(t)$ according to (\ref{equ:MAB_sample})\;
		  Reset the environment according to the setting of sampled curriculum $\Omega(t)$\;
            Record the episode of the RL agent and obtain the reward $r_i(t)$\;
            Update the policy network by (\ref{equ:ACRL})\;
            Calculate the rescaled reward $\hat{r}_i(t)$ by (\ref{Exp3:rescale})\;
            Update the corresponding weight of the target MAB $\hat{\mathrm{w}}_{i}(t+1)$ by (\ref{equ:weight_update})\;
            \If{$t\ |\ N_{MAB}=0$}
                {$\mathrm{w}_i(t)=\hat{\mathrm{w}}_i(t),\ i=0,1,...,N_{sv}^{max}$\;}
            }
             
            Output the final policy $f(\mathbf{\theta}^*)$.\
\end{algorithm}

    In our problem, the difficulty of the autonomous driving task increases with the growing number of SVs in the environment. Here, we model the self-driving tasks at unsignalized intersections with varying numbers of SVs as different arms in an MAB. An exponential algorithm is leveraged to update the importance weights associated with different arms. During the initial training phase, arms corresponding to simpler task scenarios are assigned higher importance weights than more difficult ones due to the unsatisfactory performance of the policy network and the heightened penalties incurred at dense intersections. Consequently, the RL agent undergoes more episodes in simpler task scenarios during the earlier training stages. As the policy network trains over time and achieves proficiency in simple task scenarios, it begins to allocate higher probability to the more challenging arms that are less likely to be selected before, as successful completion of tasks in more complex scenarios yields greater rewards. As the performance of the policy network improves, the MAB dynamically adjusts the weights associated with arms corresponding to different tasks, thus facilitating automated curriculum progression.

\section{Experiments}

\subsection{Comparative Experimental Settings}

The comparative experiments are conducted on the Windows 11 system with a 3.90 GHz AMD Ryzen 5 5600G CPU and an NVIDIA GeForce RTX 3060 GPU. The unsignalized intersection is constructed based on $Highway\_Env$ environment \cite{highway-env}, where each road consists of two carriageways. Here, we assume that there are up to $6$ SVs simultaneously at intersections. 
We adopt the actor-critic architecture to implement the RD-ACPPO framework. The action network and critic network are set as fully connected networks with 1 hidden layer of 128 units and 64 units by PyTorch and trained with the Adam optimizer. The number of epochs is set to 20. The learning rate of the action network and critic network are set to $5 \times 10^{-4}$ and $1 \times 10^{-3}$, respectively. $\gamma$ is set to 0.9. 

Considering the escalating difficulty of autonomous driving tasks at unsignalized intersections with the increase in the number of SVs, we empirically set the initial weights of the arms in the MAB to an exponential form that is inversely proportional to the count of SVs as follows:
\begin{equation}
\begin{split}
    \mathrm{w}_i(0)= e^{-2i},i =0,1,...,6
\end{split}
\label{MAB_exp_init}
\end{equation}

Additionally, to investigate the impact of the initial weights of the MAB arms on the training process and the performance of the trained policy, we have also conducted training for an RL policy with identical initial weights of arms in the MAB as follows:
\begin{equation}
\begin{split}
    \mathrm{w}_i(0)= 1,i =0,1,...,6
\end{split}
\label{MAB_equal_init}
\end{equation}

$\eta$ is set to 0.2. $N_{MAB}$ is set to 1000. Here, we compare the proposed RD-ACPPO method with the following baseline methods.

\begin{itemize}
    \item Fixed PPO: the agent is trained by standard PPO method \cite{schulman2017proximal} at intersections with $N_{sv}=6$.
    \item Manual CPPO: an simplified version of \cite{peng2023CPPO} with a fixed clip parameter.
    \item Random CPPO: the curriculum is sampled randomly, with an equal probability of selection for all arms.
\end{itemize}

The clip parameters of all methods were set to $\epsilon=0.2$. For the sake of fairness, other network parameters of all methods are set to be the same. The behaviors of SVs are characterized by the intelligent driver model (IDM) \cite{treiber2000congested}. 
Considering the physical limitations, the maximum acceleration and steering angle of vehicles are set to 8 m/s$^{2}$ and $45$ degrees, respectively. 
Then we test all trained policies at unsignalized intersections with $N_{sv}=0,1,...,6$ and record the success rate, collision rate, and time-out rate to evaluate the driving performance of these trained policies.

\subsection{Training Results}

\begin{figure}[t]
\centerline{\includegraphics[trim={0 0.7cm 0 0.5cm},width=0.4\textwidth]{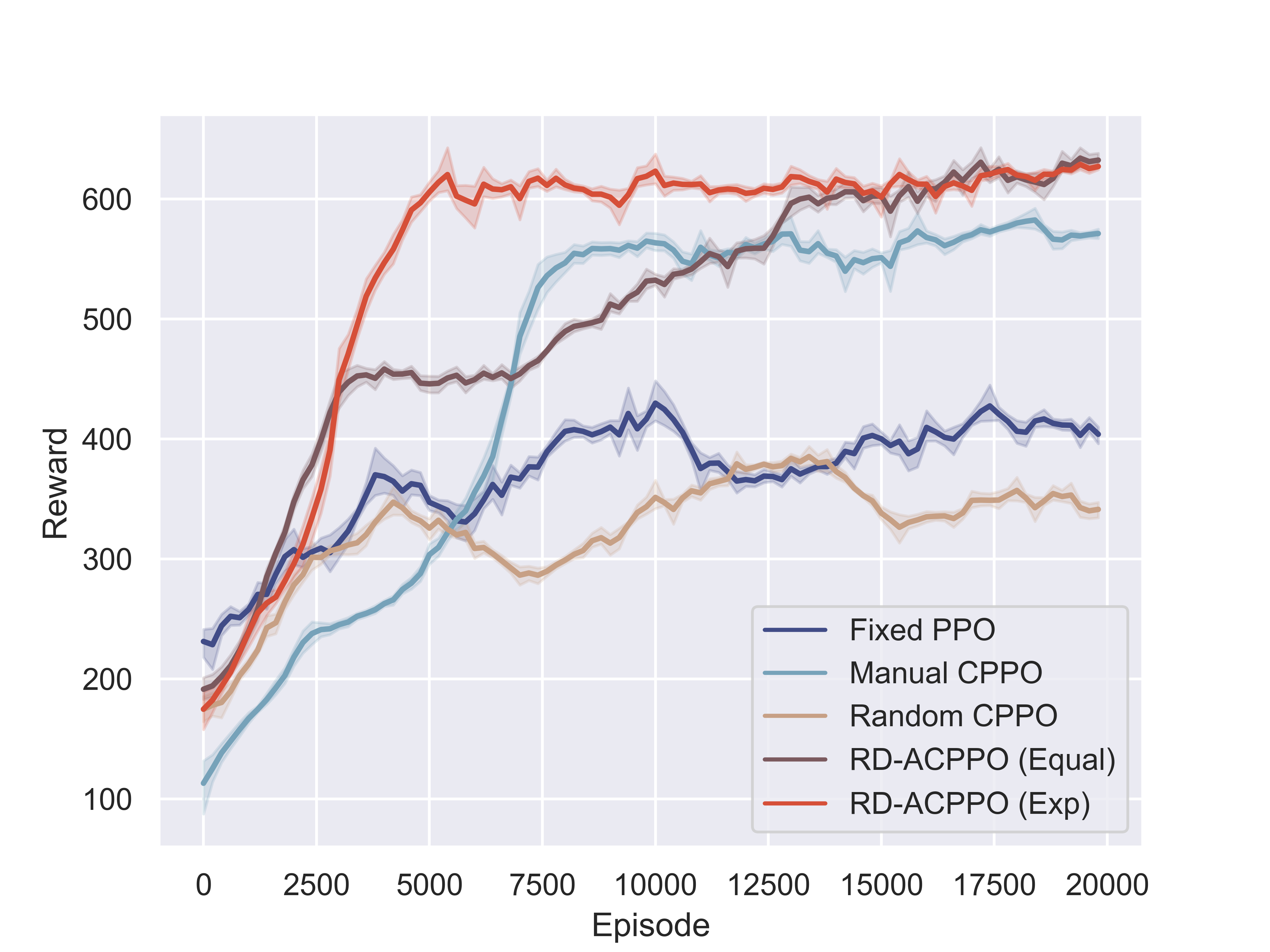}}
\caption{Reward curve comparison among different methods. The training curves are smoothed by the Savitzky-Golay filter.}
\label{reward}
\vspace{-0.2cm}
\end{figure}

\begin{figure}[t]
\centerline{\includegraphics[trim={0 0.7cm 0 0.7cm},width=0.4\textwidth]{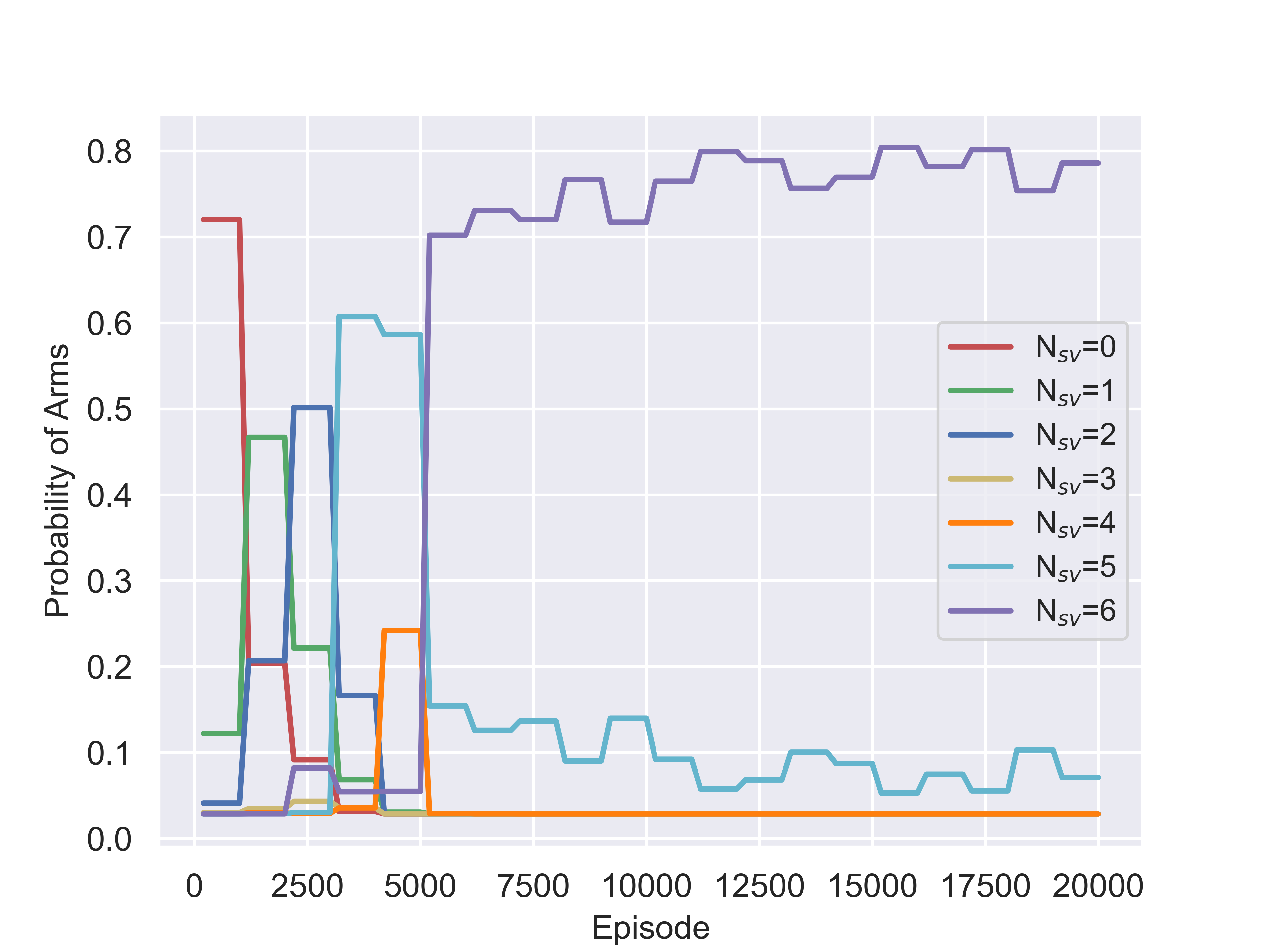}}
\caption{Probability of arms with exponential initialization weight during the training process.
}
\label{probs_exp}
\end{figure}

\begin{figure}[t]
\centerline{\includegraphics[trim={0 0.7cm 0 0.7cm},width=0.4\textwidth]{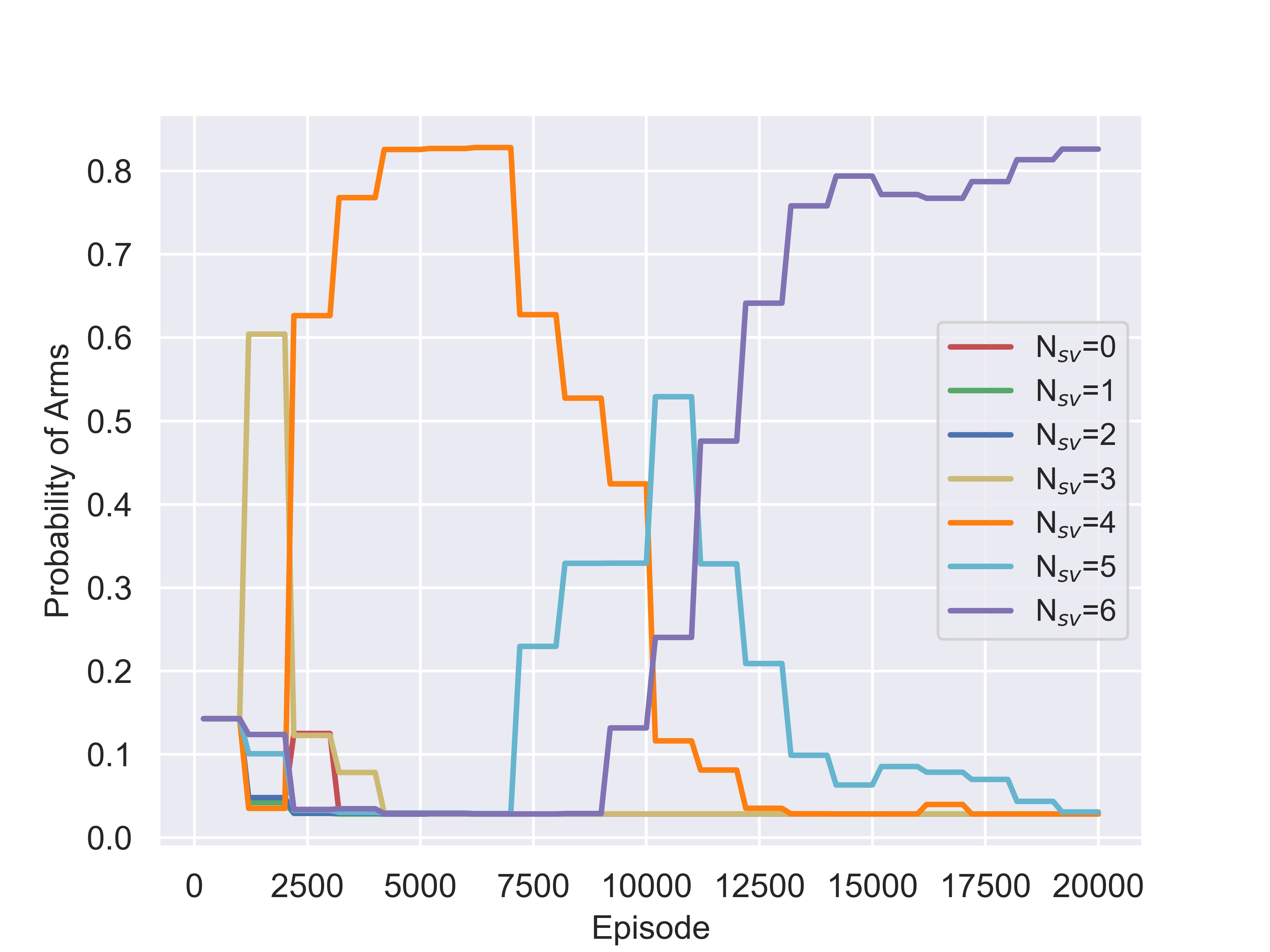}}
\caption{Probability of arms with equal initialization weight during the training process.}
\label{probs_equal}
\vspace{-0.2cm}
\end{figure}

Reward curve comparisons among different approaches and the probability changes of each arm in MAB during the training process are shown in Figs. \ref{reward}-\ref{probs_equal}. 
According to Fig. \ref{reward}, we can find that the reward curve of RD-ACPPO (Exp) initially began with lower values compared to Fixed PPO, but it surpassed that of Manual CPPO. This discrepancy can be attributed to the sensitivity of the reward function to the complexity of the training scenario, specifically characterized by the number of SVs, 
while Fixed PPO was only trained in the most complex scenarios. 
Subsequently, according to Fig. \ref{probs_exp}, due to the evaluation of potential rewards for arms of the MAB in the automated curriculum module, arms $N_{sv}=1,2,5,6$ were identified as the most preferable choices starting from the $1000$-th, $2000$-th, $3000$-th, and $5000$-th episodes in RD-ACPPO (Exp), respectively. As a result, their probability surpassed those of other arms. 
Correspondingly, the reward function of the RD-ACPPO agent experienced a rapid increase from these four time points.

It is highlighted that due to the random initialization of positions for EV and SVs, along with the sampling of arms in the MAB of the RD-ACPPO method, slight fluctuations in the reward curve still occur in the later stages of training. These fluctuations are attributed to the inherent randomness of the testing initialization, even though the RL policy has already converged.

According to Fig. \ref{probs_exp} and Fig. \ref{probs_equal}, both the MAB with equal initial weights and the MAB with exponential initial weights demonstrated a trend of curriculum selection that progressed from easier to more challenging tasks during the training process. As the RL policies were updated, there was a gradual shift towards selecting arms that offered greater potential rewards. For the MAB with exponential initial weights, in the first 3000 episodes, arms corresponding to $N_{sv}=0,1,2$ were primarily chosen. This phase allowed the learning of the capability to reach destinations and initial collision avoidance skills in environments with a few SVs at unsignalized intersections. Subsequently, in later episodes, the focus shifted to enhancing collision avoidance skills in scenarios with a high density of SVs. After the $5000$-th episode, the MAB predominantly selected the arm with $N_{sv}=6$, indicating continuous improvement in the collision avoidance capabilities with the RL policy. This enhancement enabled the RL agent to obtain pertinent rewards in the most complex unsignalized intersection environments associated with this arm, and it also facilitated the convergence of the RL policy.

According to Fig. \ref{probs_equal}, when initializing with identical weights, the weight adjustments of arms during the training process exhibited certain divergences compared to that of arms initialized with exponential weights. 
It is worth noting that the reward curve of RD-ACPPO (Equal) slightly exceeded that of RD-ACPPO (Exp) between the $1000$-th and $3000$-th episodes. 
This situation occurred because RD-ACPPO (Equal) shifted its preferable arms to $N_{sv}=3,4$, while RD-ACPPO (Exp) was still engaged in the training phase focusing on $N_{sv}=1,2$ at this stage. 
Yet, as the RL policy approached convergence, the reward curves of policies under both sets of initial conditions tended to align at a similar level. 
However, the RL policy using MABs with equal initial weights exhibited a slower convergence compared to the policy utilizing exponential initial weights. 
This result indicates that while variations in initial weight settings of MAB could influence the speed of convergence, they do not significantly affect the ultimate training outcomes of the RL policy. This resilience highlights the robustness of the proposed method, affirming its effectiveness amidst variations in initial conditions, particularly within the challenging domain of autonomous driving at unsignalized intersections.

After completing the whole training process, the reward curve of RD-ACPPO generally outperformed all baseline methods. 
This overall superiority suggests that the proposed approach has yielded the highest sample efficiency and the most effective training outcomes for the RL policy.

\subsection{Performance Evaluation}

To further evaluate the performance of the proposed RD-ACPPO algorithm, 
we conducted tests on agents trained by all methods in target scenarios with the number of SVs ranging from 0 to 6. Each agent underwent 200 repeated tests in each testing scenario, including go-straight, left-turn, and right-turn tasks. The test results are presented in Table \ref{test_result_n0} and Table \ref{test_result_more}, respectively. 
Table \ref{test_result_n0} shows that agents trained by Fixed PPO exhibit a 22.5$\%$ timeout rate in the scenario without SVs. This indicates that training the RL agent only in the scenario with $N_{sv}=6$ can lead to suboptimal policy. 
In contrast, other agents trained through curriculum learning do not exhibit this issue in this particular scenario, indicating that curriculum learning can mitigate the problem of poor generalization to empty unsignalized intersections.

\begin{figure*}[!htbp]
\centerline{\includegraphics[trim={0 0cm 0.2cm 0cm},width=0.75\textwidth]{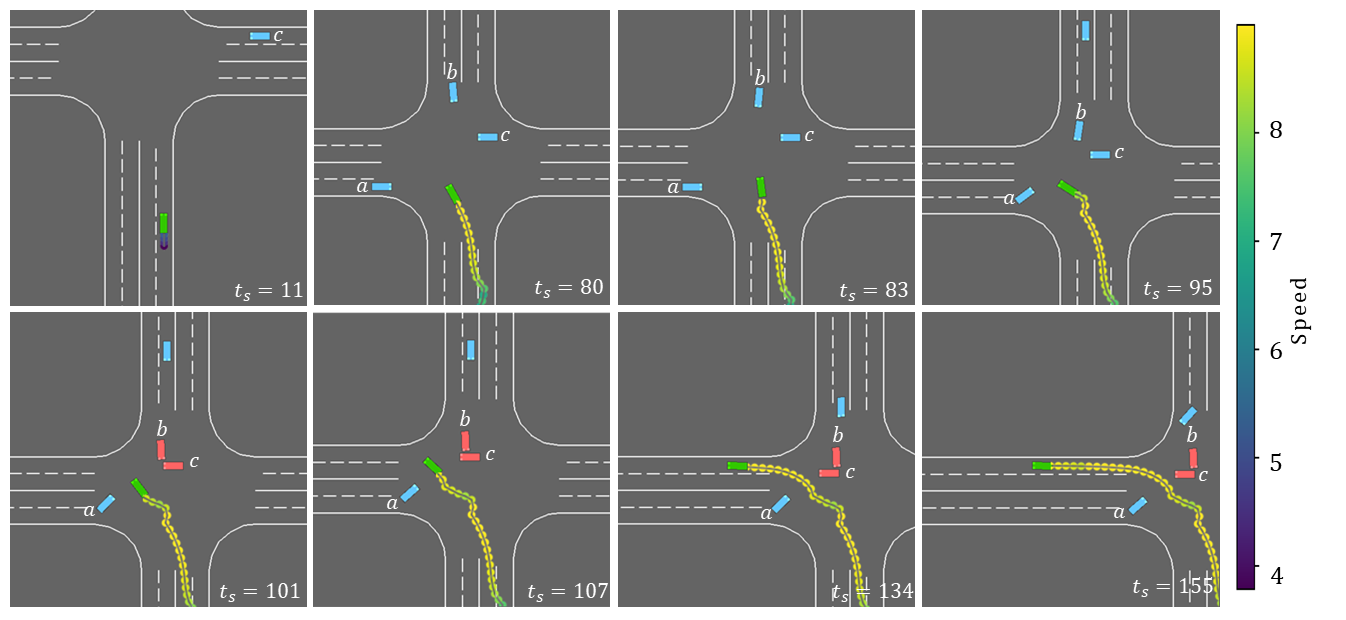}}
\caption{Demonstration of the driving performance attained by the proposed RD-ACPPO method in an unprotected left-turn task. The green car and blue cars represent the EV and SVs under normal driving conditions, respectively. The red cars represent the vehicles that have collided.
}
\label{demonstration_traj}
\vspace{-0.2cm}
\end{figure*}

\begin{figure*}[t]
\centerline{\includegraphics[trim={0cm 0.7cm 0 0.2cm},width=0.75\textwidth]{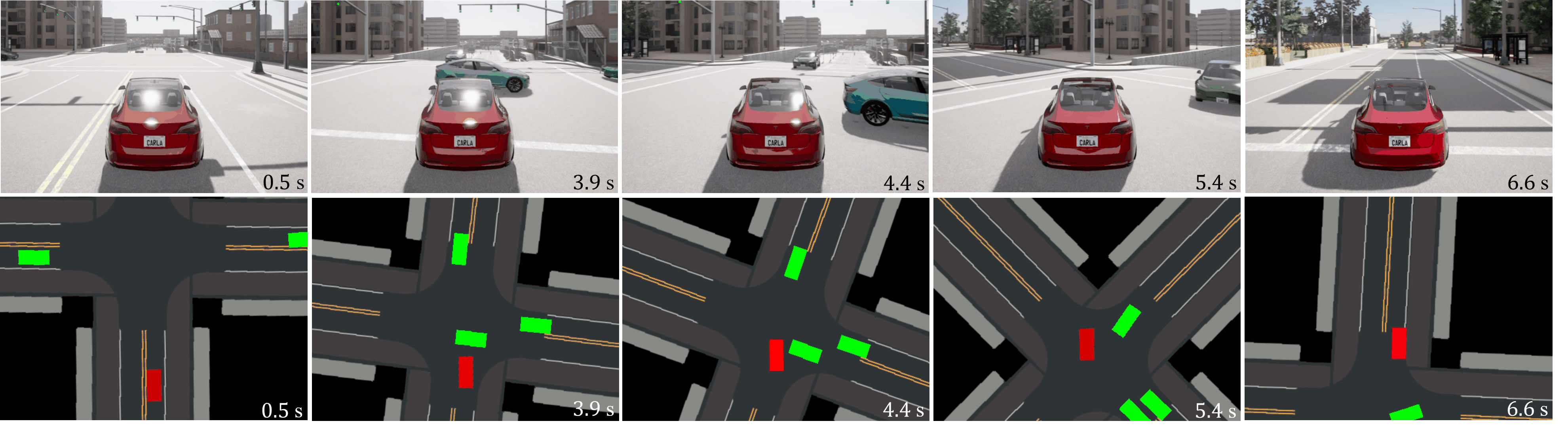}}
\caption{Key frames of the experimental validation of our method in CARLA. The top of this figure shows the third-person views of the EV, while the bottom presents the bird-eye views. 
The red rectangle and the green rectangles represent the EV and SVs, respectively.
}
\label{Carla_sim}
\vspace{-0.2cm}
\end{figure*}

\begin{table}[t]
\renewcommand{\arraystretch}{1}
\scriptsize
\centering
\caption{Performance comparison at empty unsignalized intersections among different methods.
}
\label{test_result_n0}
\begin{tabular}{c|ccc}
\hline
\multirow{2}{*}{Methods} & \multicolumn{3}{c}{$N_{sv}$=0}          \\ \cline{2-4} 
                         & succ.(\%) & coll.(\%) & time-out(\%)  \\ \hline
Fixed PPO                     & 77.5         & 0         & 22.5               \\
Manual CPPO                 & 100         & 0         & 0         \\
Random CPPO                 & 100         & 0         & 0         \\
RD-ACPPO (Equal)                & 100         & 0         & 0         \\
RD-ACPPO (Exp)                   & 100         & 0         & 0        \\ \hline
\end{tabular}
\vspace{-0.2cm}
\end{table}

\begin{table*}[t]
\renewcommand{\arraystretch}{1}
\scriptsize
\centering
\caption{Performance comparison at unsignalized intersections with different numbers of SVs among different methods.
}
\label{test_result_more}
\resizebox{\linewidth}{!}{
\begin{tabular}{c|cc|cc|cc|cc|cc|cc}
\hline
\multirow{2}{*}{Methods} & \multicolumn{2}{c|}{$N_{sv}$=1}  & \multicolumn{2}{c|}{$N_{sv}$=2}  & \multicolumn{2}{c|}{$N_{sv}$=3}  & \multicolumn{2}{c|}{$N_{sv}$=4}  & \multicolumn{2}{c|}{$N_{sv}$=5}  & \multicolumn{2}{c}{$N_{sv}$=6}    \\ \cline{2-13} 
                         & succ.(\%) & coll.(\%)  & succ.(\%) & coll.(\%)  & succ.(\%) & coll.(\%)  & succ.(\%) & coll.(\%)  & succ.(\%) & coll.(\%)  & succ.(\%) & coll.(\%)   \\ \hline
Fixed PPO                     & 92.5         & 4.5         & 83.5         & 16.5         & 75.5            & 24.5         & 73.5            & 26.5         & 67.5            & 32.5         & 64.5            & 35.5        \\
Manual CPPO                     & 99.5         & 0.5         & 87.5        & 12.5         & 80.5            & 19.5         & 76.5            & 23.5         & 73            & 27         & 70.5            & 29.5        \\
Random CPPO                     & 98         & 2         & 85.5          & 14.5         & 79            & 21         & 75            & 25         & 70           & 30         & 67.5            & 32.5        \\
\textbf{RD-ACPPO (Equal)}                     & \textbf{100}         & \textbf{0}         & \textbf{92}        & \textbf{8}         & \textbf{85.5}            & \textbf{14.5}         & \textbf{82.5}            & \textbf{17.5}         & \textbf{79}            & \textbf{21}         & \textbf{75}            & \textbf{25}        \\
\textbf{RD-ACPPO (Exp)}                     & \textbf{100}         & \textbf{0}         & \textbf{93.5}          & \textbf{6.5}         & \textbf{86.5}            & \textbf{13.5}         & \textbf{81.5}            & \textbf{18.5}         & \textbf{79}            & \textbf{21}         & \textbf{75.5}            & \textbf{24.5}       \\  \hline
\end{tabular}
}
\vspace{-0.2cm}
\end{table*}

From the testing results, we can observe that the proposed RD-ACPPO method achieved the highest task success rate across all scenarios. 
The test results of RD-ACPPO (Equal) and RD-ACPPO (Exp) across various scenario settings are similar, which is consistent with the training outcomes indicated in the reward curves.
While the success rate of RD-ACPPO decreases as the complexity of scenarios increases, it still maintains the highest success rate. This indicates that the proposed approach exhibits the best driving performance and adaptivity to environmental changes, which can be attributed to the real-time assessment provided by the automated curriculum selection mechanism during the training process. 
Among all testing results attained by the RD-ACPPO method, we pick up the results from a left-turn task for demonstration, and the details are presented in Fig. \ref{demonstration_traj}. 
The EV initiated the left-turn driving task in the lower area at the intersection and drove towards the center area. 
Then the EV decelerated and steered right when it encountered an SV (labeled as $a$) approaching from its left side in the central area. Following the right turn of the EV, two SVs (labeled as $b$ and $c$) approached the EV from the upper and from the right upper side, respectively. Simultaneously, the SV $a$ exhibited intentions of decelerating to yield. Consequently, the EV executed a left turn to avoid collisions with SV $b$ and SV $c$, subsequently realigning and continuing towards the intended lane at the top of the left area. The above actions taken by the RL agent indicated that the policy had learned to interact safely with SVs. Finally, the EV safely arrived at its destination and successfully completed the task. This example demonstrates the interaction-aware driving behavior of the RL policy trained by RD-ACPPO. 
Overall, we observe that the performance of the proposed method outperformed all baselines. This highlights the capability of the automated curriculum selection module to enhance the driving strategy performance of agents in different environments.

\subsection{Experimental Validation in CARLA}

We further validate our method in the high-fidelity simulator CARLA \cite{dosovitskiy2017carla}. The validation is conducted in the testbed of an unsignalized intersection scenario in Town03, where the maximum number of SVs is set to three. 
All vehicles on the road are set to Tesla Model 3, and SVs drive in built-in autopilot mode. Furthermore, full observability of the EV is assumed. We retrained the policy network and evaluated the driving performance in the aforementioned testbed. We test the trained policy on the unsignalized intersection and then pick up a result from a left-turn task for demonstration in CARLA. 
The key frames of a representative trail are visualized in Fig. \ref{Carla_sim}. The results illustrate that the cooperation ability of our interaction-aware driving strategy is retained. At $3.9$ s, the EV decelerated and steered towards the left side, yielding to the SV approaching from the left side. Then the EV accelerated and drove towards the target point when it observed the yield behavior of the SV coming from the upper side from $4.4$ s to $5.4$ s. Finally, the EV successfully completed a collision-free left-turn task. 
The validation results highlight the effectiveness of our approach in the high-fidelity simulator.

\section{Conclusion}

We presented the RD-ACPPO approach to train the RL policy for interaction-aware self-driving at unsignalized intersections. 
The proposed automated curriculum selection mechanism allows for the reasonable allocation of the importance weights for different curricula during various RL training periods. 
The RL policy trained by the proposed framework demonstrated reasonable curriculum transition timings during the training phase. 
The results indicated that the RD-ACPPO agent attained the highest task success rate, strong robustness to initialization parameters of MAB, and best adaptability across various test tasks. Furthermore, the effectiveness of the proposed method was validated in CARLA. 
While this work primarily focuses on the application in self-driving tasks at intersections, it can also be suitably exploited and extended to address other autonomous driving scenarios. 
\vspace{-0.3cm}

\bibliographystyle{ieeetr}
\bibliography{reference}

\begin{thebibliography}{10}

\bibitem{paden2016survey}
B.~Paden, M.~{\v{C}}{\'a}p, S.~Z. Yong, D.~Yershov, and E.~Frazzoli, ``A survey of motion planning and control techniques for self-driving urban vehicles,'' {\em IEEE Transactions on Intelligent Vehicles}, vol.~1, no.~1, pp.~33--55, 2016.

\bibitem{muhammad2020deep}
K.~Muhammad, A.~Ullah, J.~Lloret, J.~Del~Ser, and V.~H.~C. de~Albuquerque, ``Deep learning for safe autonomous driving: Current challenges and future directions,'' {\em IEEE Transactions on Intelligent Transportation Systems}, vol.~22, no.~7, pp.~4316--4336, 2020.

\bibitem{grigorescu2020survey}
S.~Grigorescu, B.~Trasnea, T.~Cocias, and G.~Macesanu, ``A survey of deep learning techniques for autonomous driving,'' {\em Journal of Field Robotics}, vol.~37, no.~3, pp.~362--386, 2020.

\bibitem{mozaffari2020deep}
S.~Mozaffari, O.~Y. Al-Jarrah, M.~Dianati, P.~Jennings, and A.~Mouzakitis, ``Deep learning-based vehicle behavior prediction for autonomous driving applications: A review,'' {\em IEEE Transactions on Intelligent Transportation Systems}, vol.~23, no.~1, pp.~33--47, 2020.

\bibitem{li2021open}
F.~Li, X.~Li, J.~Luo, S.~Fan, and H.~Zhang, ``Open-set intersection intention prediction for autonomous driving,'' in {\em Proceedings of the IEEE International Conference on Robotics and Automation (ICRA)}, pp.~13092--13098, IEEE, 2021.

\bibitem{zheng2023spatiotemporal}
L.~Zheng, R.~Yang, Z.~Peng, M.~Y. Wang, and J.~Ma, ``Spatiotemporal receding horizon control with proactive interaction towards autonomous driving in dense traffic,'' {\em arXiv preprint arXiv:2308.05929}, 2023.

\bibitem{wei2021autonomous}
L.~Wei, Z.~Li, J.~Gong, C.~Gong, and J.~Li, ``Autonomous driving strategies at intersections: Scenarios, state-of-the-art, and future outlooks,'' in {\em Proceedings of the IEEE International Intelligent Transportation Systems Conference (ITSC)}, pp.~44--51, IEEE, 2021.

\bibitem{aksjonov2021rule}
A.~Aksjonov and V.~Kyrki, ``Rule-based decision-making system for autonomous vehicles at intersections with mixed traffic environment,'' in {\em 2021 IEEE International Intelligent Transportation Systems Conference (ITSC)}, pp.~660--666, IEEE, 2021.

\bibitem{tian2020game}
R.~Tian, N.~Li, I.~Kolmanovsky, Y.~Yildiz, and A.~R. Girard, ``Game-theoretic modeling of traffic in unsignalized intersection network for autonomous vehicle control verification and validation,'' {\em IEEE Transactions on Intelligent Transportation Systems}, vol.~23, no.~3, pp.~2211--2226, 2020.

\bibitem{riegger2016centralized}
L.~Riegger, M.~Carlander, N.~Lidander, N.~Murgovski, and J.~Sj{\"o}berg, ``Centralized {MPC} for autonomous intersection crossing,'' in {\em Proceedings of the IEEE International Intelligent Transportation Systems Conference (ITSC)}, pp.~1372--1377, IEEE, 2016.

\bibitem{kneissl2018feasible}
M.~Kneissl, A.~Molin, H.~Esen, and S.~Hirche, ``A feasible {MPC}-based negotiation algorithm for automated intersection crossing,'' in {\em 2018 European Control Conference (ECC)}, pp.~1282--1288, IEEE, 2018.

\bibitem{isele2018navigating}
D.~Isele, R.~Rahimi, A.~Cosgun, K.~Subramanian, and K.~Fujimura, ``Navigating occluded intersections with autonomous vehicles using deep reinforcement learning,'' in {\em Proceedings of the IEEE International Conference on Robotics and Automation (ICRA)}, pp.~2034--2039, IEEE, 2018.

\bibitem{kiran2021deep}
B.~R. Kiran, I.~Sobh, V.~Talpaert, P.~Mannion, A.~A. Al~Sallab, S.~Yogamani, and P.~P{\'e}rez, ``Deep reinforcement learning for autonomous driving: A survey,'' {\em IEEE Transactions on Intelligent Transportation Systems}, vol.~23, no.~6, pp.~4909--4926, 2021.

\bibitem{xia2022interactive}
C.~Xia, M.~Xing, and S.~He, ``Interactive planning for autonomous driving in intersection scenarios without traffic signs,'' {\em IEEE Transactions on Intelligent Transportation Systems}, vol.~23, no.~12, pp.~24818--24828, 2022.

\bibitem{qiao2021behavior}
Z.~Qiao, J.~Schneider, and J.~M. Dolan, ``Behavior planning at urban intersections through hierarchical reinforcement learning,'' in {\em Proceedings of the IEEE International Conference on Robotics and Automation (ICRA)}, pp.~2667--2673, IEEE, 2021.

\bibitem{huang2023efficient}
Y.~Huang, S.~Yang, L.~Wang, K.~Yuan, H.~Zheng, and H.~Chen, ``An efficient self-evolution method of autonomous driving for any given algorithm,'' {\em IEEE Transactions on Intelligent Transportation Systems}, 2023.

\bibitem{bengio2009curriculum}
Y.~Bengio, J.~Louradour, R.~Collobert, and J.~Weston, ``Curriculum learning,'' in {\em Proceedings of the 26th International Conference on Machine Learning (ICML)}, pp.~41--48, 2009.

\bibitem{song2021autonomous}
Y.~Song, H.~Lin, E.~Kaufmann, P.~D{\"u}rr, and D.~Scaramuzza, ``Autonomous overtaking in gran turismo sport using curriculum reinforcement learning,'' in {\em Proceedings of the IEEE International Conference on Robotics and Automation (ICRA)}, pp.~9403--9409, IEEE, 2021.

\bibitem{wang2023chance}
Y.~Wang, Y.~Li, Z.~Peng, H.~Ghazzai, and J.~Ma, ``Chance-aware lane change with high-level model predictive control through curriculum reinforcement learning,'' {\em 2024 IEEE International Conference on Robotics and Automation (ICRA)}, 2024.

\bibitem{peng2023CPPO}
Z.~Peng, X.~Zhou, Y.~Wang, L.~Zheng, M.~Liu, and J.~Ma, ``Curriculum proximal policy optimization with stage-decaying clipping for self-driving at unsignalized intersections,'' in {\em Proceedings of the IEEE International Intelligent Transportation Systems Conference (ITSC)}, pp.~5027--5033, IEEE, 2023.

\bibitem{graves2017automated}
A.~Graves, M.~G. Bellemare, J.~Menick, R.~Munos, and K.~Kavukcuoglu, ``Automated curriculum learning for neural networks,'' in {\em Proceedings of the 34th International Conference on Machine Learning (ICML)}, pp.~1311--1320, 2017.

\bibitem{qiao2018automatically}
Z.~Qiao, K.~Muelling, J.~M. Dolan, P.~Palanisamy, and P.~Mudalige, ``Automatically generated curriculum based reinforcement learning for autonomous vehicles in urban environment,'' in {\em Proceedings of the IEEE Intelligent Vehicles Symposium (IV)}, pp.~1233--1238, IEEE, 2018.

\bibitem{khaitan2022state}
S.~Khaitan and J.~M. Dolan, ``State dropout-based curriculum reinforcement learning for self-driving at unsignalized intersections,'' in {\em Proceedings of the IEEE/RSJ International Conference on Intelligent Robots and Systems (IROS)}, pp.~12219--12224, IEEE, 2022.

\bibitem{slivkins2019introduction}
A.~Slivkins {\em et~al.}, ``Introduction to multi-armed bandits,'' {\em Foundations and Trends{\textregistered} in Machine Learning}, vol.~12, no.~1-2, pp.~1--286, 2019.

\bibitem{auer2002nonstochastic}
P.~Auer, N.~Cesa-Bianchi, Y.~Freund, and R.~E. Schapire, ``The nonstochastic multiarmed bandit problem,'' {\em SIAM Journal on Computing}, vol.~32, no.~1, pp.~48--77, 2002.

\bibitem{schulman2017proximal}
J.~Schulman, F.~Wolski, P.~Dhariwal, A.~Radford, and O.~Klimov, ``Proximal policy optimization algorithms,'' {\em arXiv preprint arXiv:1707.06347}, 2017.

\bibitem{highway-env}
E.~Leurent, ``An environment for autonomous driving decision-making.'' https://github.com/eleurent/highway-env, 2018.

\bibitem{treiber2000congested}
M.~Treiber, A.~Hennecke, and D.~Helbing, ``Congested traffic states in empirical observations and microscopic simulations,'' {\em Physical Review E}, vol.~62, no.~2, p.~1805, 2000.

\bibitem{dosovitskiy2017carla}
A.~Dosovitskiy, G.~Ros, F.~Codevilla, A.~Lopez, and V.~Koltun, ``Carla: An open urban driving simulator,'' in {\em Proceedings of the Conference on Robot Learning}, pp.~1--16, PMLR, 2017.

\end{thebibliography}
\end{document}